\documentclass[10pt,twocolumn,letterpaper]{article}

\usepackage{iccv}
\usepackage{times}
\usepackage{epsfig}
\usepackage{graphicx}
\usepackage{amsmath}
\usepackage{amssymb}
\usepackage{enumitem}
\usepackage{textcomp}

\usepackage[pagebackref=true,breaklinks=true,letterpaper=true,colorlinks,bookmarks=false]{hyperref}
\usepackage{multirow}
\usepackage{subcaption}
\usepackage{float}
\usepackage{appendix}
\usepackage{array}

\iccvfinalcopy 

\def\iccvPaperID{3105} 

\ificcvfinal\fi
\begin{document}

\title{The ``something something'' video database\\ 
for learning and evaluating visual common sense}


\author{Raghav Goyal\\
{\tt\small raghav.goyal@twentybn.com}
\and
Samira Ebrahimi Kahou\\
{\tt\small  samira.ebrahimi.kahou@gmail.com}
\and
Vincent Michalski\\
{\tt\small   michalskivince@gmail.com}
\and
Joanna Materzy\'nska\\
{\tt\small    joanna.materzynska@twentybn.com}
\and
Susanne Westphal\\
{\tt\small    susanne.westphal@twentybn.com}
\and
Heuna Kim\\
{\tt\small     heuna.kim@twentybn.com}
\and
Valentin Haenel\\
{\tt\small     valentin.haenel@twentybn.com}
\and
Ingo Fruend\\
{\tt\small     ingo.fruend@twentybn.com}
\and
Peter Yianilos\\
{\tt\small     peter@yianilos.com}
\and
Moritz Mueller-Freitag\\
{\tt\small      moritz.mueller-freitag@twentybn.com}
\and
Florian Hoppe\\
{\tt\small      florian.hoppe@twentybn.com}
\and
Christian Thurau\\
{\tt\small      christian.thurau@twentybn.com}
\and
Ingo Bax\\
{\tt\small       ingo.bax@twentybn.com}
\and
Roland Memisevic\\
{\tt\small       roland.memisevic@twentybn.com}
}

\maketitle

\begin{abstract}
Neural networks trained on datasets such as ImageNet have led to major 
advances in visual object classification. 
One obstacle that prevents networks from reasoning more deeply about complex scenes 
and situations, and from integrating visual knowledge with natural language, like humans 
do, is their lack of common sense knowledge about the physical world. 
Videos, unlike still images, contain a wealth of detailed information about the physical world. 
However, most labelled video datasets represent high-level concepts rather than detailed 
physical aspects about actions and scenes. 
In this work, we describe our ongoing collection of the ``something-something'' database 
of video prediction tasks whose solutions require a common sense understanding of the 
depicted situation. 
The database currently contains more than 100,000 videos across 174 
classes, which are defined as caption-templates. 
We also describe the challenges in crowd-sourcing this data at scale. 
\end{abstract}

\section{Introduction}
Datasets and challenges like ImageNet \cite{imagenet_cvpr09} have been major contributors  
to the recent dramatic improvements in neural network based object recognition 
\cite{krizhevsky2012imagenet,szegedy2015going,he2015deep}, 
as well as to improvements on a variety of other vision tasks thanks to transfer learning 
(eg., \cite{donahue2013decaf,sharif2014cnn,nguyen2016synthesizing}). 


\begin{figure}[]
\centering
\begin{tabular}{ccc}
\hspace{-0.3cm}
\includegraphics[width=0.14\textwidth]{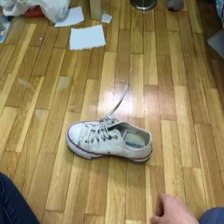}&
\includegraphics[width=0.14\textwidth]{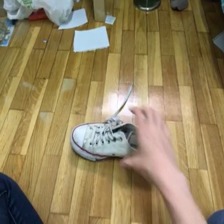}&
\includegraphics[width=0.14\textwidth]{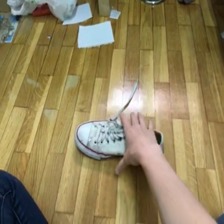}\\
\hspace{-0.3cm}
\includegraphics[width=0.14\textwidth]{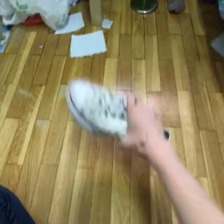}&
\includegraphics[width=0.14\textwidth]{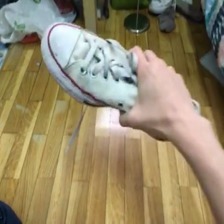}&
\includegraphics[width=0.14\textwidth]{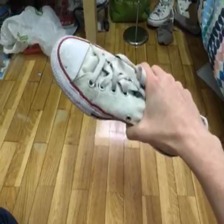}
\end{tabular}
\caption{An example video, captioned ``Picking [something] up'', taken from our growing database. 
Crowd-workers are asked to record videos and to complete caption-templates, by providing appropriate 
input-text for placeholders. 
In this example, the text provided for placeholder ``something'' is \textit{a shoe}.  
We plan to increase the complexity and sophistication of caption-templates over time, 
to the degree that models succeed at making predictions. 
}
\end{figure}

Despite their representational power, neural networks trained on still image datasets 
fall short of a variety of visual capabilities, some of which are likely 
to be resolvable using video instead of still image data. 
Specifically, 
networks trained to predict object classes from still images never observe any changes of 
objects, like changes in pose, position or distance from the observer. 
However, such changes provide important cues about object properties. 
Multiple different views of a rigid object or scene, for example, provide information 
about 3-D geometry (and it is common practice in computer vision to extract that information 
from multiple views \cite{hartley2003multiple}). 

In addition to the 3-D structure of a rigid object, the very fact that an object is rigid itself can
be extracted from multiple views of the object in motion, as well as the fact that an object is, say, articulated 
as opposed to rigid. 
Similarly, material properties such as various forms of deformability, elasticity, softness, stiffness, etc. 
express themselves visually through deformation cues encoded in multiple views.

Closely related to material properties is the notion 
of \emph{affordance} \cite{wikipedia-affordance,hofstadter2013surfaces}: An object can, by virtue of 
its properties, be used to perform certain actions, and often this usability is more important than its  
``true'' inherent object class (which may itself be considered a questionable concept \cite{hofstadter2013surfaces}). 
For example, an object that is soft and deformable, such as a blanket, can be used to cover another object; 
an object that is sharp and pointy can be used poke a hole into something; etc. 
It seems unlikely that still image tasks, which encode object properties only indirectly,  
can provide significant mileage in improving a neural network's understanding of affordances. The same 
holds true to other physical concepts that we humans intuitively grasp, like understanding that unsupported 
objects will fall (gravity) or that hidden objects do not cease to exist (object occlusion/permanence).

Motion patterns extracted from a video are not only capable of revealing object properties but also 
of revealing actions and activities. 
Not surprisingly, most of the currently popular labeled video datasets are action recognition datasets 
\cite{schuldt2004recognizing,marszalek2009actions,rodriguez2008action,karpathy2014large}.
It is important to note, however, that in a fine-grained understanding of visual concepts that goes 
beyond ``one-of-K''-labeling, actions and objects are naturally intertwined, and the tasks 
of predicting one cannot be treated independently of predicting the other. 
For example, the phrase ``opening NOUN'' will have drastically different visual 
counterparts, depending on whether ``NOUN'' in this phrase is replaced by ``door'', ``zipper'', 
``blinds'', ``bag'', or ``mouth''. There are also commonalities between these instances of ``opening'', 
like the fact that parts are moved to the sides giving way to what is behind. It is, of course, exactly these 
commonalities which define the concept of ``opening''. So a true understanding of the underlying 
meaning of the action word ``opening'' would require the ability to generalize across these different use cases. 
A proper understanding of such concepts is closely related to affordances. 
For example, the fact that a door \emph{can} be opened is much more likely to be taken into consideration, 
or even learnable by a robot (which is, say, searching for an object), if its feature space is already 
structured such that it can distinguish between opening and closing doors. 

Not only words for objects and actions can be grounded in the visual world, but also many abstract concepts, 
because these are built by means of analogy on top of the more basic, every-day concepts 
\cite{lakoff1981metaphors,hofstadter2013surfaces}.
The importance of grounding for language understanding is reflected in Winograd 
schemas \cite{levesque2011winograd}, which are linguistic puzzles whose solutions 
require common sense. A Winograd schema is given by two sentences, the second of which 
refers to the first in a superficially ambiguous way, but where the ambiguity can be 
resolved by common sense.
Consider the following Winograd schema presented in \cite{levesque2011winograd}: 
\textit{``The trophy would not fit in the brown suitcase because it was too big. What was too big?''} 
To answer this question, it is necessary to understand basic aspects of spatial relations 
and properties (like sizes), of objects, as well as the activity of putting an object into another. 
This makes it clear that text understanding from pure text may have its limitations, and 
that visual (or possibly tactile) grounding may be an inevitable step in seriously 
advancing language understanding. 




In this work, we describe our efforts in generating the ``{\it something something}''-database,
whose purpose is to provide visual (and partially acoustic) counterparts of simple, everyday aspects 
of the world. 
The goal of this data is to encourage networks to develop the features required for making 
predictions which, by definition, involve certain aspects of common sense information. 
The database\footnote{We plan to make a version of the dataset available 
at: \url{https://www.twentybn.com/datasets/something-something}}
currently contains $108,499$ \emph{short} video clips (with duration $\in [2,6]$ seconds), that are labeled 
with simple textual descriptions. The videos show objects and actions performed on them.  
Labels are in textual form and represent detailed information about the objects and actions 
as well as other relevant information. 
Predicting the textual labels from the videos requires features that are capable of representing 
physical properties of the objects and the world, such as spatial relations or material properties.

\section{Related work}
There has been an increasing interest recently in learning representations of physical aspects 
of the world using neural networks. Such representations are commonly referred to as intuitive or naive physics 
to contrast them with the symbolic/mathematical descriptions of the world developed in physics. 
Several recent papers address learning intuitive physics by using physical interactions 
(robotics) \cite{pinto2016curious,agrawal2016learning}. A shortcoming of this line of work 
is that it is based on using still images, which show, for example, how 
objects appear before and after performing a certain action. Physical predictions are made using 
convolutional networks applied to the images. Any sequential information is thus reduced to 
predicting a causal relationship between action and observations in a single 
feedforward computation, and any information encoded in the motion itself is lost. 

Some recent work has been devoted to the goal of learning about the world using video instead of robotics.
For example, there has been a long-standing endeavor to use future frames of a video as ``free'' 
labels for supervised training of neural networks. See, for example, \cite{michalski2014modeling,ranzato2014video}
and references therein. The fact that multiple images can reveal qualitatively different information 
on top of still images has been well-established for many years in both the computer vision and 
deep learning communities, and it has been exploited in the context of deep learning largely by using 
bilinear models (see \cite{memisevic2013learning} and references therein). 
Unfortunately, predicting raw pixels is challenging, both for computational and for statistical reasons. 
There are simply a lot of aspects of the real world that a predictor of raw pixels has to account for. 
This may be one reason why unsupervised learning through video prediction has hitherto 
not ``taken off''. 

One way to address the difficulty of predicting raw pixels is to re-define the 
task to predict high-level features of future frames. 
In order to obtain features one can, for example, first train a neural network on still image
object-recognition tasks \cite{vondrick2015anticipating}.  
However, a potential problem with that approach is that the purpose of using video in the first 
place is the creation of features that capture information \emph{beyond} the information already 
contained in still images. 
Also, predicting ImageNet-features has not shown yet to yield features that add substantially 
more information about the physical world. 

A hybrid between learning from video and learning from interactions is the work by \cite{lerer2016learning} 
who use a game engine to render block towers that collapse. A convolutional network is then trained to predict,  
using an image of the tower as input, whether it will collapse or not, as well as the trajectories of parts 
while the tower collapses. Similar to \cite{pinto2016curious,agrawal2016learning}, predictions are based on still 
images not videos. 

A line of work more similar to ours is Yatskar et al. \cite{yatskar2016situationrecognition}, who introduced 
a dataset of images associated with fine-grained labels by associating roles with actions and objects.  
In contrast to our work, the labels in that work describe sophisticated, cultural concepts, such 
as ``clipping a sheep's wool''. Our goal in this work is to capture basic physical concepts expressible 
in simple phrases, which we hope will provide a stepping stone towards more complex relationships and facts.
Consequently, we use natural language instead of a fixed data structure to represent labels.  
More importantly, our starting point towards fine-grained understanding is videos not images.

\section{Learning world models from video}
\label{section:background}
Although images still largely dominate research in visual deep learning, a variety of sizeable labeled 
video datasets have been introduced in recent years. The dominant application domain so far has been action 
recognition, where the task is to predict a global action label for a given video 
(for example, \cite{rodriguez2008action,karpathy2014large,marszalek2009actions,FCVID,caba2015activitynet}). 
A drawback of action recognition is that it is targeted at fairly high-level aspects of videos and 
therefore  does not encourage a network to learn about motion primitives that can encode object 
properties and intuitive physics. For example, the task associated with the datasets described in 
\cite{rodriguez2008action,karpathy2014large} is recognizing sports, and in \cite{marszalek2009actions} 
they include high-level, human-centered activities, such as ``getting out of a car'' or ``fighting''. 

A related problem is that these tasks amount to taking a long video sequence 
as input and producing one out of a relatively small number of global class-labels as the output. 
Rather than requiring a detailed understanding of what happens in a video, 
these datasets require features that can condense a long sequence (usually including 
many scene and perspective changes) into a single label. In many cases, labels can be predicted 
fairly accurately even from a single image cropped from the video. 
As a result, good classification performance can be achieved on these tasks by frame-wise aggregation 
of features extracted with a convolutional network that was pre-trained on a \emph{still image} task, 
such as ImageNet \cite{wu2016harnessing}. 
This is in stark contrast to the goal of using video in order to learn a better 
world model. 
One prerequisite for learning more fine-grained information about the world 
is that labels describe video content that is \emph{restricted to a short time interval}.  
Only this way can there be a tight synchronization between video content and the corresponding labels,
allowing learned features to correspond to physical aspects of the unfolding scene, 
and to correlate with the low-level aspects that are reflected, for example, also in everyday language. 

Detailed labeling has been addressed also in various \emph{video captioning} datasets recently, 
where the goal is to predict an elaborate description, rather than a single label, 
for a video \cite{torabi2015using,rohrbach2015dataset,xu2016msr, krishna2017dense}. 
However, similar to many of the action recognition datasets mentioned above, they typically contain 
descriptions that reflect high-level, cultural aspects of human life and commonly require a 
good knowledge of rare or unusual facts and language.  
Furthermore, since descriptions summarize fairly long videos, they do not have a high temporal resolution.

\begin{table*}[]
\centering
\caption{Comparison with other video datasets recorded specifically for training machine learning models (information partially taken from \cite{krishna2017dense}).}
\label{tab:comparison-dataset}
\begin{tabular}{|l|l|l|l|l|}
\hline
\textbf{Dataset}           & \textbf{Domain}       & \textbf{\# Videos} & \textbf{Avg. duration} & \textbf{Remarks}                                                    \\ \hline
Physics 101 \cite{phys101}                & intuitive physics     & 17,408             & -                      & \begin{tabular}[c]{@{}l@{}}101 objects with 4 different \\ scenarios (ramp, spring, fall, liquid)\end{tabular} \\
MPII cooking \cite{rohrbach2012database}               & action (cooking)      & 44                 & 600s                   & -                                                                     \\
TACoS \cite{regneri2013grounding}                     & action (cooking)      & 127                & 360s                   & -                                                                     \\
Charades \cite{sigurdsson2016hollywood}                   & action (human)        & 10, 000            & 30s                    & -                                                                     \\
KITTI \cite{geiger2013vision}                     & action (driving)      & 21                 & 30s                    & -                                                                     \\
Something-Something (ours) & \begin{tabular}[c]{@{}l@{}}human-object \\  interaction\end{tabular} & 108,499           & 4.03s                  & \begin{tabular}[c]{@{}l@{}}174 fine-grained categories of \\ human-object interaction  scenarios\end{tabular}             \\ \hline
\end{tabular}
\end{table*}

A dataset focussing on lower-level, more physical concepts is described in 
\cite{wu2016computational}. 
The dataset contains $17,408$ videos of a small set of objects involved in a number 
of physical experiments. These include, for example, letting the object slide down a slope or 
dropping it onto a surface. The supervision signal is given by (known) physical properties 
of the experiment, such as the angle of the slope or the material of the object. 
In contrast to that work, besides scaling to a much larger size, we use language as labels, 
similar to captioning datasets. 
This allows us to generate a much larger and more varied set of actions and labels. 
It also allows us to go beyond a small and highly specialized set of physical properties 
and actions prescribed by the experimental setup and by what can easily be measured. 


Many shortcomings of existing video datasets may be related to the fact that 
they are generated by annotating (or using closed captionings of) existing video material, 
including excerpts from Hollywood movies. 
Recently, \cite{sigurdsson2016hollywood} proposed a way to overcome this problem by asking 
crowdworkers to record videos themselves rather than to attach labels to existing videos. 
In this work, we follow a similar approach using a scalable framework for crowd-sourced 
video recording. Using our large-scale crowd acting\textsuperscript{TM} framework, we have 
so far generated several hundred thousand videos, including the dataset discussed in this paper. 
In contrast to the dataset described in \cite{sigurdsson2016hollywood} we focus here 
on basic, physical concepts 
rather than on higher-level human activities. 
A comparison with existing similar datasets is shown in Table~\ref{tab:comparison-dataset}


\section{The ``something-something'' dataset}
In this work, we introduce the ``something-something''-dataset.  
It currently contains $108,499$ videos across $174$ labels, with duration 
ranging from 2 to 6 seconds. 
Labels are textual descriptions based on templates, such 
as ``Dropping [{\it something}] into [{\it something}]'' containing slots 
(``[{\it something}]'') that serve as placeholders for objects. 
Crowd-workers provide videos where they act out the templates. They 
choose the objects to perform the actions on and enter the noun-phrase describing the 
objects when uploading the videos. 

The dataset is split into train, validation and test-sets in the ratio of 8:1:1.
The splits were created so as to ensure that all videos provided by the same worker 
occur only in one split (train, validation, or test). 
See Table~\ref{tab:data_specs} for some summary information about the dataset. 

\begin{table}[]
\centering
\begin{tabular}{|l|l|}
\hline
\multicolumn{2}{|c|}{Dataset Specifications} \\ \hline
Number of videos                     & 108,499 \\
Number of class labels               & 174   \\
Average duration of videos (in seconds) & 4.03  \\
Average number of videos per class  & 620 \\
\hline
\end{tabular}
\caption{Dataset summary}
\label{tab:data_specs}
\end{table}

\begin{figure*}[]
\centering
\includegraphics[width=1.0\textwidth]{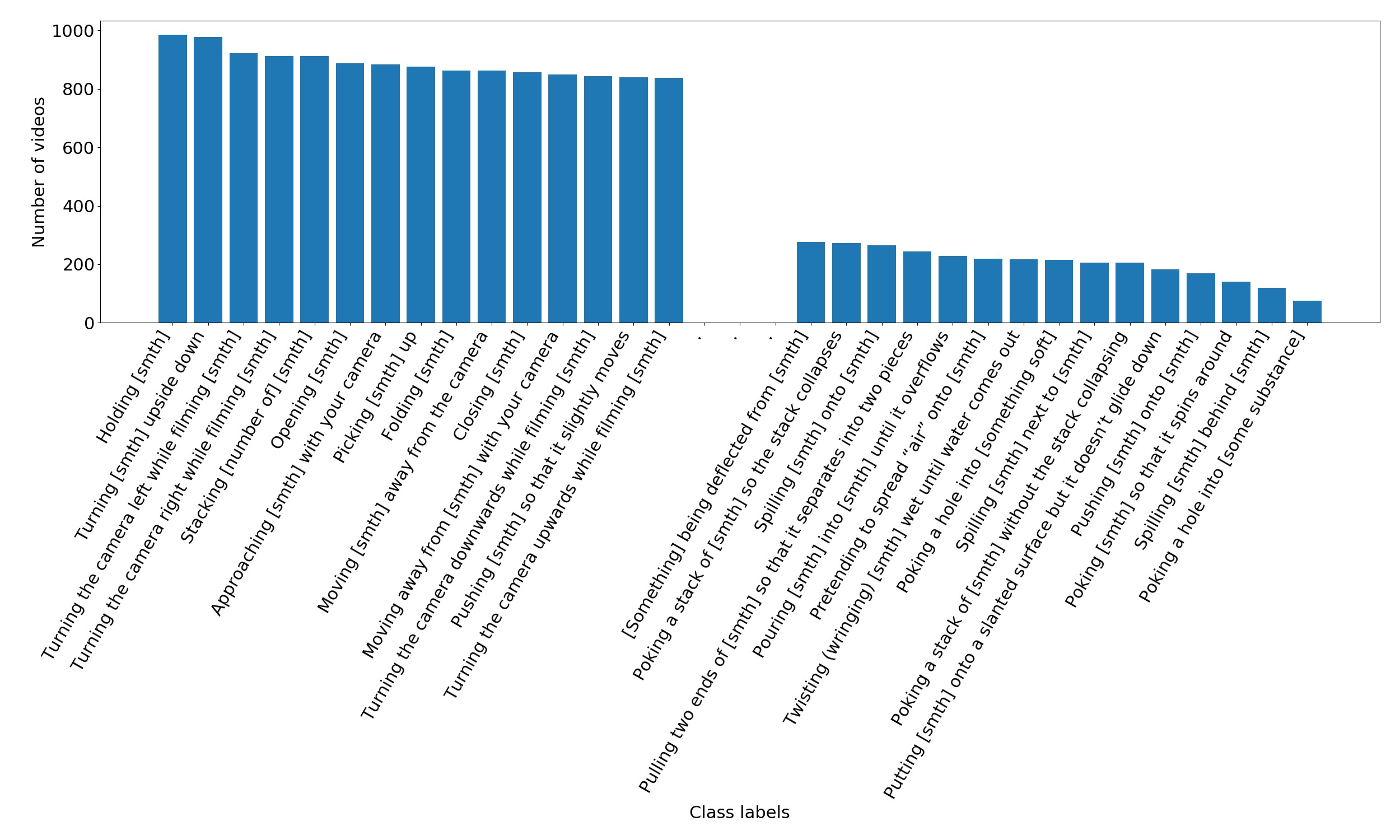}
\caption{Numbers of videos per class (truncated for better visualisation).}
\label{figure:class_frequencies}
\end{figure*}

\begin{figure*}
\centering
\begin{subfigure}{.5\textwidth}
  \centering
  \includegraphics[width=.9\linewidth]{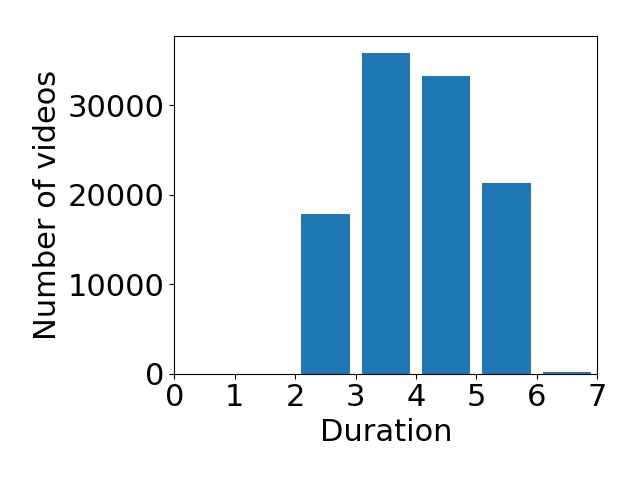}
  \caption{Video lengths (in seconds), ranging between 2 and 6 seconds (inclusive)}
  \label{figure:duration_hist}
\end{subfigure}%
\begin{subfigure}{.5\textwidth}
  \centering
  \includegraphics[width=1.0\linewidth]{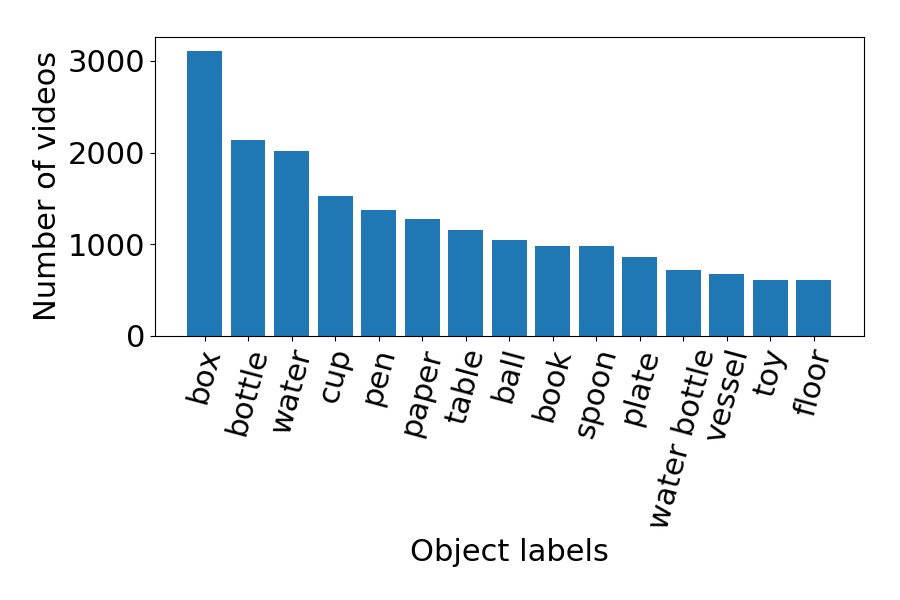}
  \caption{Frequencies of occurrence of 15 most common objects}
  \label{figure:object_frequencies}
\end{subfigure}
\end{figure*}

Including differences in case, stemming, use of determiners, etc., $23,137$ distinct 
object names have been submitted in the current version of the dataset. 
We estimate the number of actually distinct objects to be at least a few thousand. 
Figure~\ref{figure:object_frequencies} shows the frequency of objects for the most common objects.

\begin{figure*}[h]
\centering
\begin{tabular}{c@{\hskip 0.1em} c@{\hskip 0.1em} c@{\hskip 0.1em} c@{\hskip 0.1em} c@{\hskip 0.1em} c@{\hskip 0.1em}}

\includegraphics[width=0.162\textwidth]{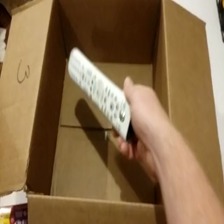}&
\includegraphics[width=0.162\textwidth]{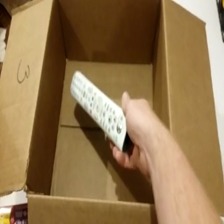}&
\includegraphics[width=0.162\textwidth]{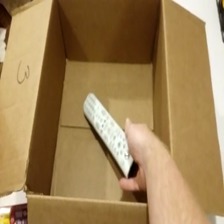}&
\includegraphics[width=0.162\textwidth]{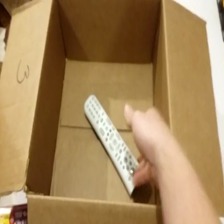}&
\includegraphics[width=0.162\textwidth]{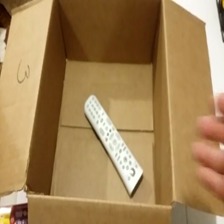}&
\includegraphics[width=0.162\textwidth]{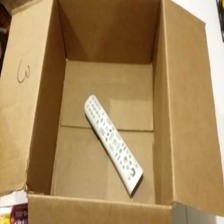}\\
\multicolumn{6}{c}{Putting \emph{a white remote} into \emph{a cardboard box}}\\

\includegraphics[width=0.162\textwidth]{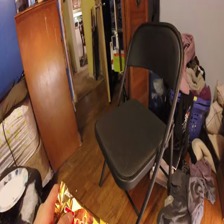}&
\includegraphics[width=0.162\textwidth]{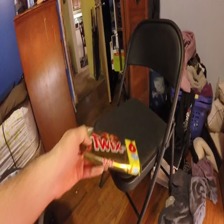}&
\includegraphics[width=0.162\textwidth]{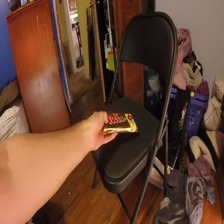}&
\includegraphics[width=0.162\textwidth]{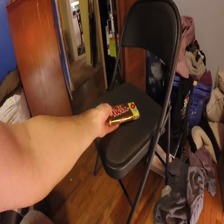}&
\includegraphics[width=0.162\textwidth]{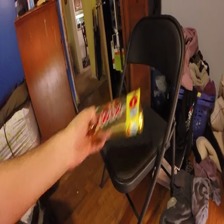}&
\includegraphics[width=0.162\textwidth]{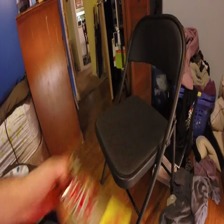}\\
\multicolumn{6}{c}{Pretending to put \emph{candy} onto \emph{chair}}\\

\includegraphics[width=0.162\textwidth]{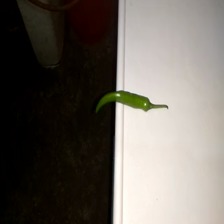}&
\includegraphics[width=0.162\textwidth]{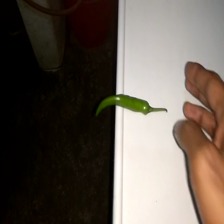}&
\includegraphics[width=0.162\textwidth]{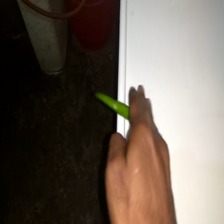}&
\includegraphics[width=0.162\textwidth]{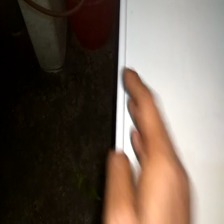}&
\includegraphics[width=0.162\textwidth]{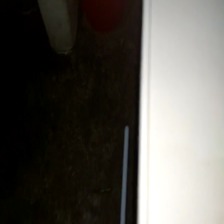}&
\includegraphics[width=0.162\textwidth]{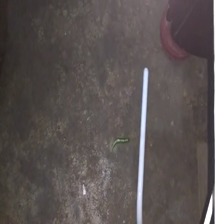}\\
\multicolumn{6}{c}{Pushing \emph{a green chilli} so that it falls off the table}\\

\includegraphics[width=0.162\textwidth]{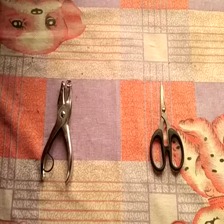}&
\includegraphics[width=0.162\textwidth]{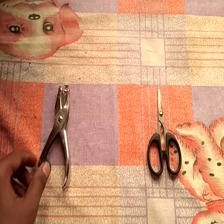}&
\includegraphics[width=0.162\textwidth]{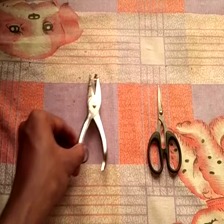}&
\includegraphics[width=0.162\textwidth]{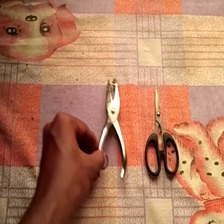}&
\includegraphics[width=0.162\textwidth]{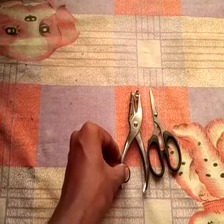}&
\includegraphics[width=0.162\textwidth]{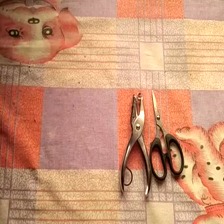}\\
\multicolumn{6}{c}{Moving \emph{puncher} closer to \emph{scissor}}\\

\end{tabular}
\caption{Example videos and corresponding descriptions. Object entries shown in italics.}
\label{figure:examplevideos}
\end{figure*}

In its current version, the dataset was generated by $1133$ crowd workers with an average of $127.32$ workers per class.
Figure~\ref{figure:class_frequencies} shows a truncated distribution of the number of videos per class, 
with an average of roughly $620$ videos per class, a minimum of $77$ for ``Poking a hole into [\textit{some substance}]'' 
and a maximum of $986$ for ``Holding [\textit{something}]''.
Figure~\ref{figure:duration_hist} shows a histogram of the duration of videos (in seconds).  
A few examples of frame samples from the collected videos is shown in Figure \ref{figure:examplevideos}. 

\subsection{Crowdsourced video recording}
The currently pre-dominant way of creating large, labeled datasets is to start by gathering 
a large collection of input items, such as images or videos. Usually, these are found 
using online resources, such as Google image search or Youtube. Subsequently, 
the gathered input examples are labeled using crowdsourcing services like Amazon Mechanical 
Turk (AMT) (see, for example, \cite{imagenet_cvpr09}). 

As outlined is Section~\ref{section:background} videos available online are largely unsuitable 
for the goal of learning simple (but fine-grained) visual concepts. 
We therefore ask crowd-workers to provide videos \emph{given} labels instead of the other way 
around (a similar approach was recently described in \cite{sigurdsson2016hollywood}). 

\subsection{Natural language labels and curriculum learning}
\label{subsection:descriptions}

The number of ``everyday concepts'' that we want to capture with this dataset is gigantic, 
and it cannot be captured within a fixed set of one-hot labels. 
Natural language descriptions are a natural and obvious solution to this problem: 
natural language is capable of representing an extremely large number of ``classes'' 
and it is compositional and thereby able to express this number highly economically. 

Unfortunately, natural language provides a much weaker learning signal than 
a one-hot label. This is one reason why image and video captioning systems are currently 
trained using an ImageNet pre-trained network as the vision component. 

To obtain useful natural-language labels, but also be able to train, and potentially bootstrap, 
networks to learn from the data, we generate natural language descriptions automatically by 
appropriately combining partly pre-defined, and partly human-generated, parts of speech. 
Natural language descriptions take the form of \emph{templates} that AMT workers provide 
along with videos, as we shall describe in the next section. 
Analogous to how probabilistic graphical models impose independence assumptions on a 
multivariate distribution, these ``structured captions'', can be viewed as approximations 
to full natural language descriptions, that allow us to control the complexity of learning 
by imposing a rich (but restricted) structure on the labels. 

In the current version of the dataset, we emphasize short and simple descriptions, most of 
which contain only the most important parts of speech, such as verbs, nouns and prepositions. 
This choice was made, because common neural networks are not yet able to represent elaborate 
captions and high-level concepts. 

However, it is possible to increase the degree of complexity as well as the sophistication of 
language over time as the dataset grows. This approach can be viewed as ``curriculum learning'' 
\cite{bengio2009curriculum}, where simple concepts are taught first, and more complicated 
concepts are added progressively over time. 
From this perspective, the level of complexity of the current version of the dataset may be viewed 
approximately as ``teaching a one-year-old child''. Unlike labels that are encoded using a fixed datatype, 
as described, for example, in \cite{yatskar2016situationrecognition}, natural language labels allow us to represent 
a spectrum of complexity, from simple objects and actions encoded as one-hot labels, to full-fledged captions.
The use of natural language encodings for classes furthermore allows us to dynamically adjust the label 
structure in response to how learning progresses. In other words, the complexity of videos and natural 
language descriptions can be increased as a function of the validation-accuracy achievable by networks trained 
on the data so far. 



\subsection{Non-uniform sampling of the Cartesian product of actions and objects}
\label{subsection:cartesianproduct}
Although it is more restricted than captions, the cross product of actions and objects constitutes 
a space that is so large that there is no hope to sample it sufficiently densely as needed for practical applications. 
But the empirical probability density of real-world cases in the space of permissible actions and objects is far from 
uniform. Many actions, such as ``Moving an elephant on the table'' or ``Pouring paper from a cup'', 
for example, have almost zero density. And more reasonable combinations can still have highly variable 
probabilities. Consider, for example, ``drinking from a plastic bag'' (highly rare) vs. ``dropping a 
piece of paper'' (highly common). 

It is possible to exploit the low entropy of this distribution, by using the following 
sampling scheme: Each crowd-worker is presented with an action in the form of a template that contains 
one or several placeholders for objects.
Workers then get to decide which objects to perform the action on and generate a video clip. 
When uploading the video, workers are required to enter their object choice(s) into a provided mask. 

\subsection{Grouping and contrastive examples} 
\label{subsection:contrastiveexamples}
The goal of the ``something-something'' collection effort is to provide fine-grained discrimination tasks, 
whose solution will require a fairly deep understanding of the physical world. 
However, especially in the early stage, where simple descriptions focussed on verbs and nouns dominate, 
networks can learn to ``cheat'', for example, by extracting the object type from one or several 
individual frames, and by extracting the action using indirect cues, such as hand 
position, overall velocity, camera shake, etc. 
This is an example of \emph{dataset bias} \cite{torralba2011unbiased}. 

As a way to reduce bias, by forcing networks to classify the actual actions and the underlying physics, 
we provide \emph{action groups} for most action types. 
An action group contains multiple similar actions with minor visual differences,  
so that fine-grained understanding of the activity is required to distinguish the actions within a group. 
Providing action groups to the AMT workers also encourages these to perform the multiple 
different actions with the same object, such that a close attention to detail is required 
to correctly identify the action within the group. We found that action groups also serve 
the communication with crowdworkers in clarifying to them the kinds of fine-grained distinctions 
in the uploaded videos we expect. 

Some groups contain \emph{pretending} actions in addition to the actual action to be 
performed. This will require any system training on this data to closely observe the object 
instead of secondary cues such as hand positions. It will also require the networks to learn and represent 
indirect visual cues, such as the fact that an object is present or not present in a particular region in the image. 
Preventing a network from ``cheating'' by distinguishing between actual and pretended actions 
is reminiscent of teaching a child by asking it to tell the difference between genuine and false actions. 
Examples of action groups we use include: 
\begin{itemize}[noitemsep]
   \setlength\itemsep{0.1em}
    \item Putting \emph{something} on top of \emph{something} / Putting \emph{something} next to \emph{something} / Putting \emph{something} behind \emph{something} 
    \item Putting \emph{something} behind \emph{something} / Pretending to put \emph{something} behind \emph{something} (but not actually leaving it there)
    \item Poking \emph{something} so lightly that it does not or almost does not move / Poking \emph{something} so it slightly moves / Poking \emph{something} so that it falls over.
    \item Poking \emph{something} / Pretending to poke \emph{something} 
\end{itemize}
A more comprehensive list of action groups and descriptions examples are 
provided in the supplementary materials. 

%
%
%
%

\subsection{Data collection platform}
\label{subsection:platform}
Besides the requirements outlined above, crowdsourcing the recording of video data according 
to a pre-defined label structure poses a variety of technical challenges: 
\begin{itemize}[noitemsep]
    \item \emph{Batch submission:} Crowd workers need to be able to initiate a job, and come back to it later potentially multiple times until it is completed, so that they can record videos outside or at other places or times of the 
day, or after having gathered the objects needed for the task. 
    \item \emph{Worker-conditional choice of labels:} To generate data with sufficient variability, it is important that each label is represented by videos from as many different crowdworkers as possible. To this end, it is necessary to keep track of the set of labels recorded by each individual crowdworker. `The list of labels or action groups (as defined below) to choose from can be generated dynamically once the crowdworker logs on to the platform. 
    \item \emph{Feedback on completed or partially completed submissions:} In the case of submissions that are fully or partially rejected it is important that the crowd sourcing operators can quickly provide feedback to the crowd workers regarding what was wrong with the submission. 
    \item \emph{Convenience:} To reduce cost, crowd workers need to face a convenient, easy-to-use and highly responsive interface. 
\end{itemize}
To address these challenges, we created a data collection platform, with 
which both crowd workers and our operators overseeing the crowdsourcing efforts 
interact during the ongoing crowdsourcing operation. 

When an AMT crowdworker accepts a task on the AMT interface he/she gets re-directed 
to our platform, where the task is then completed and reviewed. After completion 
of a task, our platform communicates with AMT to communicate the result (accept/reject) 
and allow for payments for the accepted tasks.

On the platform, workers get presented with a list of action-templates 
to choose from (with action-templates grouped as described in the previous section). 
By selecting action-templates, the platform creates video upload-boxes where 
workers can upload the videos as required, along with label-templates with variable-roles 
to be filled by workers.
After uploading a video, all variable-roles in the label template (represented by the 
word ``something'' in most of our label templates) turn into input masks, and the worker is 
asked to fill in the correct word (such as the noun describing the object used).
Each uploaded video is displayed (as screenshot) in a video playback-box and it can be played back 
for easy inspection by the workers (as well as by the operators as we describe below). 
After the worker reaches the number of requested videos, a button ``Submit Hit'' gets released, 
that allows the worker to submit the assignment and get paid. 

A submission is accepted automatically, if it passes a number of quality control checks, 
which verify aspects such as length and uniqueness of the videos. 
Every submission is subsequently verified for correctness by a human operator. 
For more details on the crowd acting platform and screenshots we refer to the supplementary materials.

\section{Baseline experiments}
We performed a few baseline experiments to assess the difficulty of the task of 
predicting label templates from the videos. 
In this work, we discuss classification tasks on the label templates. Full captioning and 
performance on the expanded labels will be discussed elsewhere. 
On the classification tasks, we found 3d-convolutional networks to generally outperform 
2d-convolutional networks and their combination to work best. 
But we also found that many of the subtle classes that were chosen explicitly to make the 
task harder (Section~\ref{subsection:contrastiveexamples}), 
are hardly distinguishable using these fairly standard architectures. 
More sophisticated architectures are necessary to obtain better performance on this data. 
A difficulty for both training and interpreting results is the presence of ambiguities in the labels.  
For reporting, these can be dealt with to some degree by resorting to top-K error rate. 
Both ambiguities and the overall difficulty of the prediction tasks can be alleviated 
by choosing label subsets and by combining labels into groups, 
which can allow fairly simple architectures to achieve reasonable performance. 
We shall discuss several such simplified subsets of classes below. 
We also found that this grouping can help as an initialization for networks that are 
subsequently fine-tuned on more complex class-choices.  

\subsection{Pre-processing}
For the baseline runs, we sample frames from the videos using a frame rate of $24$ fps and resize 
them to a resolution of $84 \times 84$ pixels, 
except for those runs where we use a pre-trained model (in which case we use the resolution is determined by that model).
We lowpass-filter the resulting videos in time using a Gaussian kernel with zero mean and variance 
of $48$ pixels, which was chosen to largely eliminate frequencies above the Nyquist-frequency, 
taking into consideration the target frame-rate of $6$ frames per second (as discussed below).

We also perform temporal augmentation by choosing a random offset between $0$ and the downsampling 
factor ($4$) during training. We use a fixed offset of $0$ for validation and testing. 
We have also experimented with other types of data augmentation including flipping frames for invariant 
classes and random rotation by a small angle, but we did not find any significant performance gains for these.

\begin{table}[]
\centering
\begin{tabular}{|c|}
\hline
\textbf{10 selected classes}                               \\ \hline
Dropping {[}something{]} \\
Moving {[}something{]} from right to left \\
Moving {[}something{]} from left to right \\
Picking {[}something{]} up \\
Putting {[}something{]} \\
Poking {[}something{]} \\
Tearing {[}something{]} \\
Pouring {[}something{]} \\
Holding {[}something{]} \\
Showing {[}something{]} (almost no hand) \\ \hline
\end{tabular}
\caption{Subset of $10$ hand-chosen ``easy'' classes.}
\label{tab:10-selected-classes}
\end{table}

\begin{table*}[]
\centering
\begin{tabular}{|c|c|c|c|c|c|c|c|}
\hline
\multirow{3}{*}{\textbf{Method}} & \multicolumn{7}{c|}{\textbf{Error rate (\%)}}                                                                  \\ \cline{2-8} 
                                 & \multicolumn{2}{c|}{10 classes} & \multicolumn{2}{c|}{40 classes} & \multicolumn{3}{c|}{174 classes}           \\ \cline{2-8} 
                                 & top-1          & top-2          & top-1          & top-2          &    top-1       &      top-2     &  top-5   \\ \hline\hline
2D CNN + Avg                     & 76.5           & 58.9           & 88.0           & 78.5           & -                & -               & -          \\ \hline
Pre-2D CNN + Avg                 & 54.7           & 39.0           & 79.2           & 70.0           & -                & -                & -          \\ \hline
Pre-2D CNN + LSTM                & 52.3           & 34.1           & 77.8           & 68.0           & -                & -                & -          \\ \hline\hline
3D CNN + Stack                   & 58.1           & 38.7           & 70.3           & 57.3           & -                & -                & -          \\ \hline
Pre-3D CNN + Avg                 & 47.5           & 29.2           & 66.2           & 52.7           &    88.5        &     81.5       &  70.0    \\ \hline\hline
2D+3D-CNN                        & \bf{44.9}      & \bf{27.1}      & \bf{63.8}      & \bf{50.7}      & -                & -                & -          \\ \hline
\end{tabular}
\caption{Error rates on different subsets of the data.}
\label{tab:results-table}
\end{table*}

\subsection{Model specifications}
Here we report results on the task of predicting action templates using multiple different encoding methods. 
We found dropout on the first fully-connected layer and batch-normalization on the last layer 
to significantly improve training. The encoding methods we used are: 

\textbf{2D-CNN + Avg:} Using the VGG-16 net architecture \cite{simonyan2014very} to represent individual frames and averaging the obtained features for each frame in the video to form the final encoding. The weights of the network were trained from scratch.

\textbf{Pre-2D-CNN + Avg:} Using an Imagenet-trained VGG-16 architecture to represent individual frames and averaging the obtained features for each frame in the video to form the final encoding. 

\textbf{Pre-2D-CNN + LSTM:} Using the above pre-trained VGG network to represent individual frames and passing the extracted features to an LSTM layer with a hidden state size of $256$. The last hidden state of the LSTM is then taken as the video encoding. 

\textbf{3D-CNN + Stack:} Using a 3D-CNN model trained from scratch with specifications following \cite{tran2015learning}, but with a size of $1024$ units for the fully-connected layers and a clip size of $9$ frames. We extract these features from non-overlapping clips of size $9$ frames (after padding all videos to a maximal length of $36$ frames), and stack the obtained features to obtain a $4096$ dimensional representation ($4$ columns), masking the column features, such that invalid frames (due to padding) do not affect training.

\textbf{Pre-3D-CNN + Avg:} Using a 3D-CNN model initialized on the sports-1m dataset \cite{tran2015learning} and finetuned on our dataset. In this case, we use the framerate $8$ fps for training and extract columns of size $16$ frames with $8$ frames overlap between columns, such that the total number of columns is $5$. 
We average the features across the clips.

\textbf{2D+3D-CNN:} A combination of the best performing 2D-CNN and 3D-CNN trained models, obtained by concatenating the two resulting video-encodings.

\subsection{Results}

\begin{figure*}[h]
\begin{tabular}{cccccl}
\includegraphics[width=0.12\textwidth]{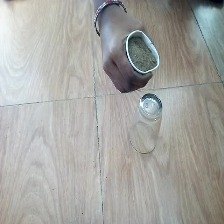}&
\includegraphics[width=0.12\textwidth]{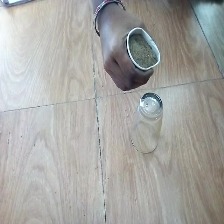}&
\includegraphics[width=0.12\textwidth]{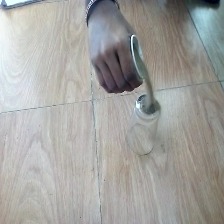}&
\includegraphics[width=0.12\textwidth]{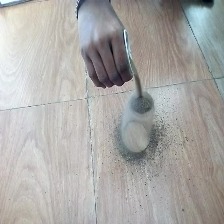}&
\includegraphics[width=0.12\textwidth]{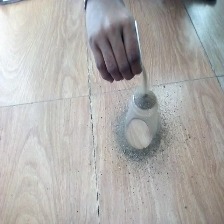}&
\begin{tabular}[b]{@{}l@{}}
1. Pouring [something]: 77.47\%\\
2. Putting [something]: 8.48\% \\ \\ \\
\end{tabular}\\
\multicolumn{5}{c}{Original: Pouring [something]}\\

\includegraphics[width=0.12\textwidth]{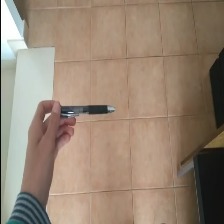}&
\includegraphics[width=0.12\textwidth]{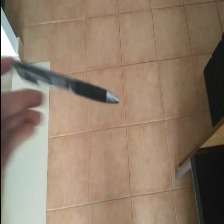}&
\includegraphics[width=0.12\textwidth]{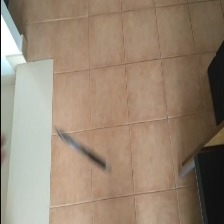}&
\includegraphics[width=0.12\textwidth]{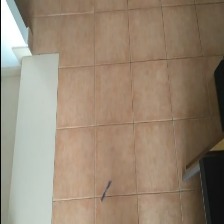}&
\includegraphics[width=0.12\textwidth]{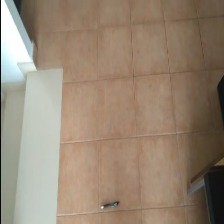}&
\begin{tabular}[b]{@{}l@{}}
1. Dropping [something]: 65.6\%\\
2. Poking [something]: 14.38\% \\ \\ \\
\end{tabular}\\
\multicolumn{5}{c}{Original: Dropping [something]}\\

\includegraphics[width=0.12\textwidth]{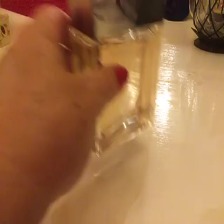}&
\includegraphics[width=0.12\textwidth]{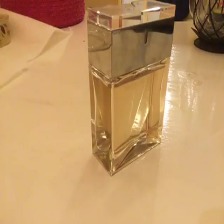}&
\includegraphics[width=0.12\textwidth]{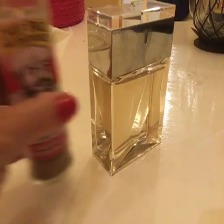}&
\includegraphics[width=0.12\textwidth]{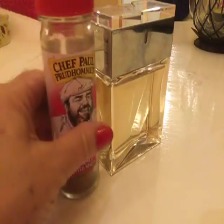}&
\includegraphics[width=0.12\textwidth]{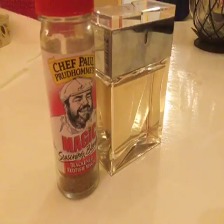}&
\begin{tabular}[b]{@{}l@{}}
1. Putting [something]: 97.41\%\\
2. Picking [something] up: 0.72\% \\ \\ \\
\end{tabular}\\
\multicolumn{5}{c}{Original: Putting [something]}\\

\includegraphics[width=0.12\textwidth]{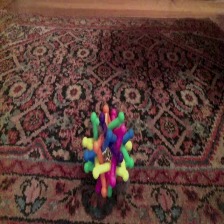}&
\includegraphics[width=0.12\textwidth]{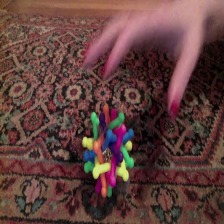}&
\includegraphics[width=0.12\textwidth]{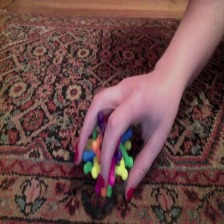}&
\includegraphics[width=0.12\textwidth]{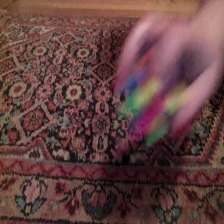}&
\includegraphics[width=0.12\textwidth]{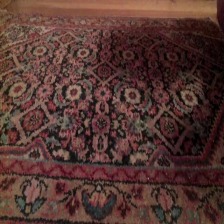}&
\begin{tabular}[b]{@{}l@{}}
1. Picking [something] up: 62.07\%\\
2. Putting [something]: 29.52\% \\ \\ \\
\end{tabular}\\
\multicolumn{5}{c}{Original: Picking [something] up}\\

\includegraphics[width=0.12\textwidth]{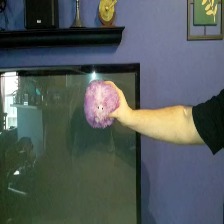}&
\includegraphics[width=0.12\textwidth]{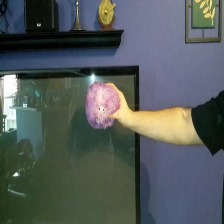}&
\includegraphics[width=0.12\textwidth]{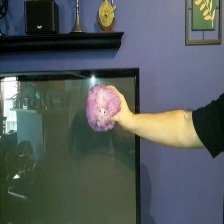}&
\includegraphics[width=0.12\textwidth]{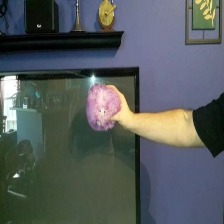}&
\includegraphics[width=0.12\textwidth]{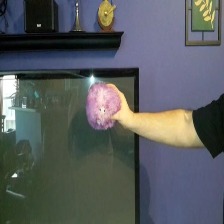}&
\begin{tabular}[b]{@{}l@{}}
1. Showing [something]\\(almost no hand): 69.64\%\\
2. Holding [something]: 15.19\% \\ \\ \\
\end{tabular}\\
\multicolumn{5}{c}{Original: Holding [something]}\\

\includegraphics[width=0.12\textwidth]{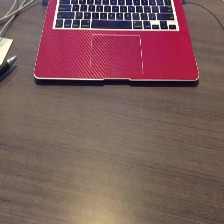}&
\includegraphics[width=0.12\textwidth]{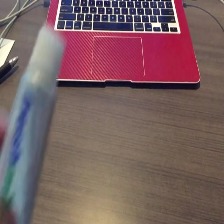}&
\includegraphics[width=0.12\textwidth]{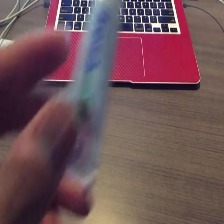}&
\includegraphics[width=0.12\textwidth]{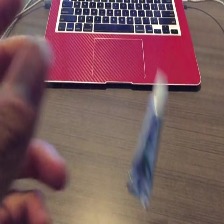}&
\includegraphics[width=0.12\textwidth]{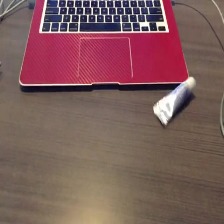}&
\begin{tabular}[b]{@{}l@{}}
1. Putting [something]: 53.69\%\\
2. Dropping [something]: 40.28\% \\ \\ \\
\end{tabular}\\
\multicolumn{5}{c}{Original: Dropping [something]}\\

\includegraphics[width=0.12\textwidth]{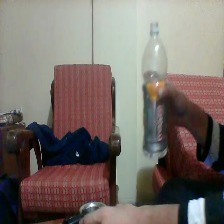}&
\includegraphics[width=0.12\textwidth]{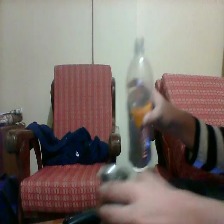}&
\includegraphics[width=0.12\textwidth]{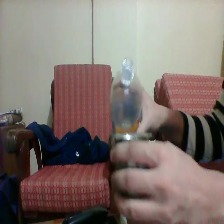}&
\includegraphics[width=0.12\textwidth]{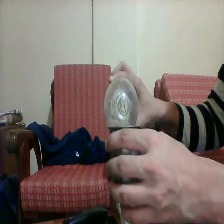}&
\includegraphics[width=0.12\textwidth]{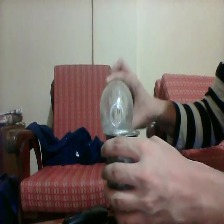}&
\begin{tabular}[b]{@{}l@{}}
1. Picking [something] up: 65.58\%\\
2. Putting [something]: 11.41\% \\ \\ \\
\end{tabular}\\
\multicolumn{5}{c}{Original: Pouring [something]}\\

\includegraphics[width=0.12\textwidth]{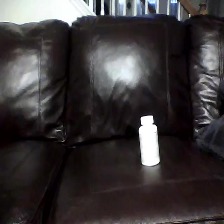}&
\includegraphics[width=0.12\textwidth]{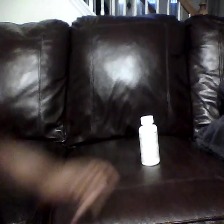}&
\includegraphics[width=0.12\textwidth]{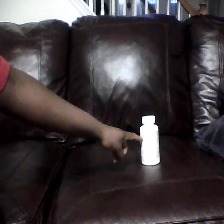}&
\includegraphics[width=0.12\textwidth]{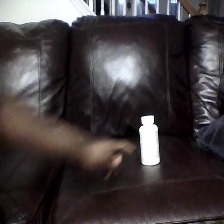}&
\includegraphics[width=0.12\textwidth]{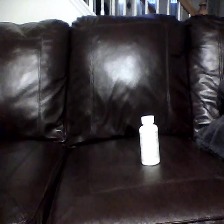}&
\begin{tabular}[b]{@{}l@{}}
1. Picking [something] up: 72.49\%\\
2. Putting [something]: 25.07\% \\ \\ \\
\end{tabular}\\
\multicolumn{5}{c}{Original: Poking [something]}\\

\end{tabular}
\caption{Some example predictions from the  best performing baseline model on 10 selected classes experiment}
\label{figure:examplepredictions}
\end{figure*}

We compared these networks mainly on two subsets of the dataset with classes hand-picked to simplify 
the task and benchmark  
the complexity of the dataset (we refer to the supplementary materials for more details on selection of classes):
\textbf{10 selected classes:} We first pre-select $41$ ``easy'' classes. We then generate $10$ classes to train the networks (shown in Table~\ref{tab:10-selected-classes}), where each class is formed by grouping together one or more of the original $41$ classes with similar semantics. The mapping from $41$ to $10$ classes is shown in Table~\ref{tab:map-10-classes} in the appendix. The total number of videos in this case is $28198$.  
\textbf{40 selected classes:} Keeping the above $10$ groups, we select $30$ additional common classes. The total number of samples in this case is $53267$.
Some example predictions from the $10$-class model are shown in Figure~\ref{figure:examplepredictions}.

We show the error rates for these subsets using the baselines described above in Table~\ref{tab:results-table}.
It shows that the difficulty of the task grows significantly as the number of classes are 
increased (despite the corresponding growth of the training-set). 
Similar to datasets like Imagenet, ambiguities in the labels make the naive classification performance 
look deceptively weak. 
However, even the top-2 performance shows that there the dataset poses a significant challenge 
for these architectures. 

We also experimented on \textbf{all 174 classes} using a 3D CNN model pre-trained on the $40$ selected classes, 
and obtained error rates of top-1: $88.5\%$, top-5: $70.3\%$. 

Overall, our results demonstrate that the presence of subtle distinctions (using grouping, contrastive 
examples and other design choices) makes this an extraordinarily difficult problem for standard architectures. 

\section{Discussion}
Advances in common sense reasoning can come mainly from two sources: 
through learning from interactions with the world, and through learning from observing the world. 
The first, interactions, rely crucially on advances in robotics. 
Unlike human interactions, however, robotic interactions lack the sophisticated 
tactile sensing that allows, for example, blind humans to learn about the world 
without any vision. 
It seems likely that even a robotics-based approach to learning common 
sense will rely on highly capable visual perception and on visuomotor policies that 
can deal with video input. 

The dataset and learning methods that we describe in this work fall into the second 
category: learning about the world through vision. 
In contrast to unsupervised approaches, based on video-prediction, 
we propose approaching the problem through supervised learning on fine-grained labeling tasks. 


The database introduced in this paper is an ongoing collection effort. 
We will continue to grow and extend the dataset over time in response to, and as a function of,  
the ability of networks to learn from this data. 








{\small
\bibliographystyle{ieee}
\bibliography{ms}
}
	
\newpage
\appendix

\section{Dataset}

\subsection{10-selected classes}
The mapping used for generating the classes used in the $10$-class experiments is shown 
in Table \ref{tab:map-10-classes}. The $10$ classes were defined by first selecting $41$ classes 
by hand (based on class-definitions and on visual inspection of the videos) and subsequently remapping these 
into $10$ groups. 


\begin{figure*}[h]
    \centering
    \includegraphics[width=1.1\textwidth]{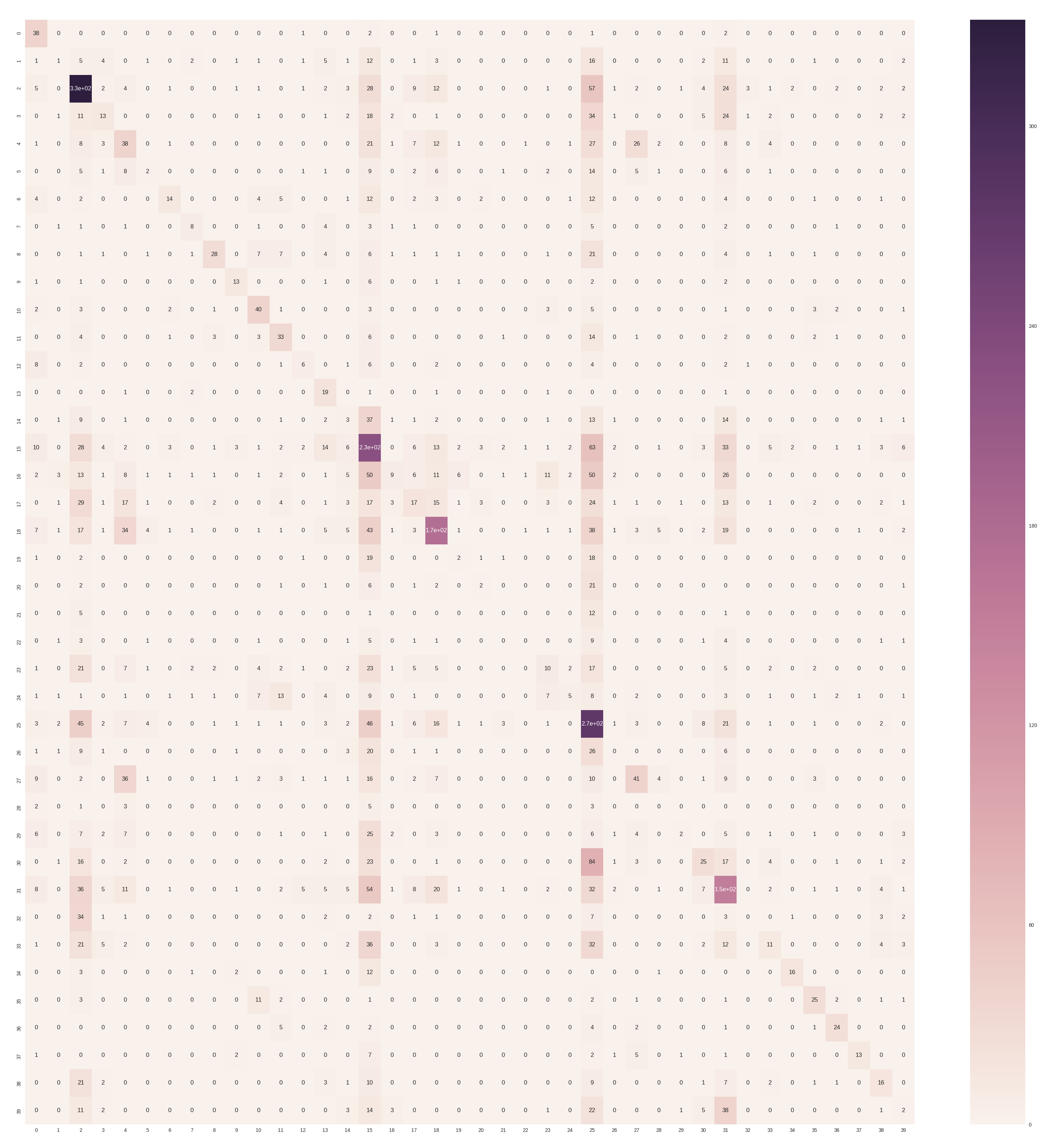}
    \caption{Confusion matrix for the best model trained on the $40$ selected classes. Corresponding class-names are listed in Table~\ref{tab:40-classes-list}.}
\label{fig:conf-mat-40-classes}
\end{figure*}

\begin{table}[h]
\centering
\begin{tabular}{|c|c|}
\hline
\textbf{Class names}                      & \textbf{Index} \\ \hline
Dropping {[}something{]}                  & 0              \\
Holding {[}something{]}                   & 1              \\
Moving {[}something{]} from left to right & 2              \\
Moving {[}something{]} from right to left & 3              \\
Picking {[}something{]} up                & 4              \\
Poking {[}something{]}                    & 5              \\
Pouring {[}something{]}                   & 6              \\
Putting {[}something{]}                   & 7              \\
Showing {[}something{]} (almost no hand): & 8              \\
Tearing {[}something{]}                   & 9              \\ \hline
\end{tabular}
\caption{Subset of 10 selected classes used in some of the experiments.}
\label{tab:10-classes-dict}
\end{table}

\subsection{40-selected classes}
We took the above 10 selected classes and select 30 additional common classes to form this subset of data. The list of classes used is shown in the Table \ref{tab:40-classes-list}.

The confusion matrix for predictions on 40 selected classes using the best performing model is shown in Figure \ref{fig:conf-mat-40-classes} and the corresponding dictionary to read classes off from the matrix is shown in the Table \ref{tab:40-classes-list}.

\begin{table}[]
\centering
\begin{tabular}{|l|l|}
\hline
\multicolumn{1}{|c|}{\textbf{Class names}}           & \multicolumn{1}{c|}{\textbf{Index}} \\ \hline
Approaching {[}something{]} with your camera               & 0                                   \\
Closing {[}something{]}                                    & 1                                   \\
Dropping {[}something{]}                                   & 2                                   \\
Folding {[}something{]}                                    & 3                                   \\
Holding {[}something{]}                                    & 4                                   \\
Holding {[}something{]} next to {[}something{]}            & 5                                   \\
Moving {[}something{]} away from {[}something{]}           & 6                                   \\
Moving {[}something{]} away from the camera                & 7                                   \\
Moving {[}something{]} closer to {[}something{]}           & 8                                   \\
Moving {[}something{]} down                                & 9                                   \\
Moving {[}something{]} from left to right                  & 10                                  \\
Moving {[}something{]} from right to left                  & 11                                  \\
Moving {[}something{]} towards the camera                  & 12                                  \\
Moving away from {[}something{]} with your camera          & 13                                  \\
Opening {[}something{]}                                    & 14                                  \\
Picking {[}something{]} up                                 & 15                                  \\
Plugging {[}something{]} into {[}something{]}              & 16                                  \\
Poking {[}something{]}                                     & 17                                  \\
Pouring {[}something{]}                                    & 18                                  \\
Pretending to pick {[}something{]} up                      & 19                                  \\
Pretending to put {[}something{]} next to {[}something{]}  & 20                                  \\
Pretending to put {[}something{]} on a surface             & 21                                  \\
Pretending to take {[}something{]} from {[}somewhere{]}    & 22                                  \\
Pushing {[}something{]} so that it slightly moves          & 23                                  \\
Pushing {[}something{]} with {[}something{]}               & 24                                  \\
Putting {[}something{]}                                    & 25                                  \\
Putting {[}something{]} into {[}something{]}               & 26                                  \\
Showing {[}something{]} (almost no hand)                   & 27                                  \\
Showing a photo of {[}something{]} to the camera           & 28                                  \\
Showing that {[}something{]} is empty                      & 29                                  \\
Stacking {[}number of{]} {[}something{]}                   & 30                                  \\
Tearing {[}something{]}                                    & 31                                  \\
Throwing {[}something{]} against {[}something{]}           & 32                                  \\
Turning {[}something{]} upside down                        & 33                                  \\
\begin{tabular}[c]{@{}l@{}}Turning the camera downwards while\\ filming {[}something{]}\end{tabular} & 34
\\
Turning the camera left while filming {[}something{]}      & 35                                  \\
Turning the camera right while filming {[}something{]}     & 36                                  \\
\begin{tabular}[c]{@{}l@{}}Turning the camera upwards while\\ filming {[}something{]}\end{tabular} & 37
\\
Uncovering {[}something{]}                                 & 38                                  \\
Unfolding {[}something{]}                                  & 39                                  \\ \hline
\end{tabular}
\caption{Subset of 40 selected classes used in some of the experiments.}
\label{tab:40-classes-list}
\end{table}

\begin{table*}[]
\centering
\caption{Mapping used for 10 selected classes}
\begin{tabular}{|l|l|}
\hline
\multicolumn{1}{|c|}{\textbf{Actual class}}                                              & \multicolumn{1}{c|}{\textbf{Mapped class}}                 \\ \hline
{[}Something{]} falling like a rock                                                      & \multirow{5}{*}{Dropping {[}something{]}}                  \\
{[}Something{]} falling like a feather or paper                                          &                                                            \\
Throwing {[}something{]}                                                                 &                                                            \\
Throwing {[}something{]} onto a surface                                                  &                                                            \\
Throwing {[}something{]} in the air and letting it fall                                  &                                                            \\ \hline
Pushing {[}something{]} from right to left                                               & \multirow{2}{*}{Moving {[}something{]} from right to left} \\
Pulling {[}something{]} from right to left                                               &                                                            \\ \hline
Pulling {[}something{]} from left to right                                               & \multirow{2}{*}{Moving {[}something{]} from left to right} \\
Pushing {[}something{]} from left to right                                               &                                                            \\ \hline
Picking {[}something{]} up                                                               & \multirow{7}{*}{Picking {[}something{]} up}                \\
Lifting {[}something{]} up completely without letting it drop down                       &                                                            \\
Moving {[}something{]} up                                                                &                                                            \\
Lifting {[}something{]} with {[}something{]} on it                                       &                                                            \\
Taking {[}something{]} from {[}somewhere{]}                                              &                                                            \\
Taking {[}one of many similar things on the table{]}                                     &                                                            \\
Taking {[}something{]} out of {[}something{]}                                            &                                                            \\ \hline
Putting {[}something{]} next to {[}something{]}                                          & \multirow{9}{*}{Putting {[}something{]}}                   \\
Putting {[}something{]} onto {[}something{]}                                             &                                                            \\
Putting {[}something{]} on a surface                                                     &                                                            \\
Putting {[}something similar to other things that are already on the table{]}            &                                                            \\
Putting {[}something{]} behind {[}something{]}                                           &                                                            \\
Putting {[}something{]}, {[}something{]} and {[}something{]} on the table              &                                                            \\
Putting {[}something{]} and {[}something{]} on the table                                 &                                                            \\
Putting {[}something{]} on a flat surface without letting it roll                        &                                                            \\
Putting {[}something{]} that can't roll onto a slanted surface, so it stays where it is  &                                                            \\ \hline
Poking {[}something{]} so that it falls over                                             & \multirow{4}{*}{Poking {[}something{]}}                    \\
Poking {[}something{]} so lightly that it doesn't or almost doesn't move                 &                                                            \\
Poking a stack of {[}something{]} so the stack collapses                                 &                                                            \\
Poking a stack of {[}something{]} without the stack collapsing                           &                                                            \\ \hline
Tearing {[}something{]} into two pieces                                                  & \multirow{2}{*}{Tearing {[}something{]}}                   \\
Tearing {[}something{]} just a little bit                                                &                                                            \\ \hline
Pouring {[}something{]} into {[}something{]}                                             & \multirow{5}{*}{Pouring {[}something{]}}                   \\
Pouring {[}something{]} onto {[}something{]}                                             &                                                            \\
Pouring {[}something{]} out of {[}something{]}                                           &                                                            \\
Pouring {[}something{]} into {[}something{]} until it overflows                          &                                                            \\
Trying to pour {[}something{]} into {[}something{]}, but missing so it spills next to it &                                                            \\ \hline
Holding {[}something{]}                                                                  & \multirow{2}{*}{Holding {[}something{]}}                   \\
Holding {[}something{]} in front of {[}something{]}                                      &                                                            \\ \hline
Showing {[}something{]} on top of {[}something{]}                                        & \multirow{3}{*}{Showing {[}something{]} (almost no hand)}  \\
Showing {[}something{]} behind {[}something{]}                                           &                                                            \\
Showing {[}something{]} next to {[}something{]}                                          &                                                            \\ \hline
\end{tabular}
\label{tab:map-10-classes}

\end{table*}

\subsection{All data - 175 classes}
The complete list of $175$ classes and their corresponding action-groups is shown 
in Table~\ref{tab:all_classes}.

\onecolumn
\begin{table}[H]
\centering
\caption{Class labels and their corresponding action-groups for all 175 classes} 
\label{tab:all_classes}
\begin{tabular}{|p{0.65\textwidth}|p{0.35\textwidth}|}
\hline
\multicolumn{1}{|c|}{\textbf{Class Labels}}                                                                                                         & \multicolumn{1}{c|}{\textbf{Action Groups}}                                  \\ \hline
\begin{tabular}[c]{@{}l@{}}Trying but failing to attach {[}something{]} to {[}something{]} because it doesn't stick\end{tabular}                 & \multirow{2}{*}{Attaching/Trying to attach}                                  \\
Attaching {[}something{]} to {[}something{]}                                                                                                        &                                                                              \\ \hline
Bending {[}something{]} until it breaks                                                                                                             & \multirow{3}{*}{Bending something}                                           \\
Trying to bend {[}something unbendable{]} so nothing happens                                                                                        &                                                                              \\
Bending {[}something{]} so that it deforms                                                                                                          &                                                                              \\ \hline
Digging {[}something{]} out of {[}something{]}                                                                                                      & \multirow{2}{*}{Burying or digging something}                                \\
Burying {[}something{]} in {[}something{]}                                                                                                          &                                                                              \\ \hline
Moving away from {[}something{]} with your camera                                                                                                                                                                                                          & \multirow{8}{*}{Camera motions}                                              \\
Turning the camera right while filming {[}something{]}                                                                                                                                                                                                                                                                                                           &                                                                              \\
Approaching {[}something{]} with your camera                                                                                                        &                                                                              \\
Turning the camera left while filming {[}something{]}                                                                                                                                                                                                                                                                                                            &                                                                              \\
Turning the camera upwards while filming {[}something{]}                                                                                                                                                                                                                                                                                                         &                                                                              \\
Moving {[}something{]} away from the camera                                                                                                                                                                                                                &                                                                              \\
Moving {[}something{]} towards the camera                                                                                                                                                                                                                  &                                                                              \\
Turning the camera downwards while filming {[}something{]}                                                                                                                                                                                                                                                                                                       &                                                                              \\ \hline
{[}Something{]} colliding with {[}something{]} and both are being deflected                                                                         & \multirow{3}{*}{Collisions of objects}                                       \\
{[}Something{]} being deflected from {[}something{]}                                                                                                &                                                                              \\
{[}Something{]} colliding with {[}something{]} and both come to a halt                                                                              &                                                                              \\ \hline
Uncovering {[}something{]}                                                                                                                          & \multirow{2}{*}{Covering}                                                    \\
Covering {[}something{]} with {[}something{]}                                                                                                       &                                                                              \\ \hline
Putting {[}something similar to other things that are already on the table{]}                                                                       & \multirow{2}{*}{Crowd of things}                                             \\
Taking {[}one of many similar things on the table{]}                                                                                                &                                                                              \\ \hline
Dropping {[}something{]} into {[}something{]}                                                                                                                                                                                                              & \multirow{5}{*}{Dropping something}                                          \\
Dropping {[}something{]} onto {[}something{]}                                                                                                       &                                                                              \\
Dropping {[}something{]} next to {[}something{]}                                                                                                    &                                                                              \\
Dropping {[}something{]} in front of {[}something{]}                                                                                                &                                                                              \\
Dropping {[}something{]} behind {[}something{]}                                                                                                     &                                                                              \\ \hline
Showing {[}something{]} next to {[}something{]}                                                                                                                                                                                                                                                                                                                  & \multirow{3}{*}{Filming objects, without any actions}                        \\
Showing {[}something{]} on top of {[}something{]}                                                                                                                                                                                                                                                                                                                &                                                                              \\
Showing {[}something{]} behind {[}something{]}                                                                                                                                                                                                                                                                                                                   &                                                                              \\ \hline
Folding {[}something{]}                                                                                                                             & \multirow{2}{*}{Folding something}                                           \\
Unfolding {[}something{]}                                                                                                                           &                                                                              \\ \hline
Hitting {[}something{]} with {[}something{]}                                                                                                        & Hitting something with something                                             \\ \hline
Holding {[}something{]} in front of {[}something{]}                                                                                                 & \multirow{5}{*}{Holding something}                                           \\
Holding {[}something{]} behind {[}something{]}                                                                                                      &                                                                              \\
Holding {[}something{]} next to {[}something{]}                                                                                                                                                                                                            &                                                                              \\
Holding {[}something{]}                                                                                                                                                                                                                                    &                                                                              \\
Holding {[}something{]} over {[}something{]}                                                                                                        &                                                                              \\ \hline
Lifting up one end of {[}something{]}, then letting it drop down                                                                                                                                                                                           & \multirow{4}{*}{Lifting and (not) dropping something}                        \\
Lifting up one end of {[}something{]} without letting it drop down                                                                                                                                                                                         &                                                                              \\
Lifting {[}something{]} up completely, then letting it drop down                                                                                                                                                                                           &                                                                              \\
Lifting {[}something{]} up completely without letting it drop down                                                                                                                                                                                         &                                                                              \\ \hline
Tilting {[}something{]} with {[}something{]} on it until it falls off                                                                               &   \multirow{3}{*}{\parbox{6cm}{Lifting/Tilting objects with other objects on them}}          \\
Lifting {[}something{]} with {[}something{]} on it                                                                                                  &                                                                              \\
Tilting {[}something{]} with {[}something{]} on it slightly so it doesn't fall down                                                                 &                                                                              \\ \hline
Moving {[}something{]} up                                                                                                                                                                                                                                  & \multirow{2}{*}{Moving something}                                            \\
Moving {[}something{]} down                                                                                                                                                                                                                                &                                                                              \\ \hline
\end{tabular}
\end{table}
\begin{table}[H]
\begin{tabular}{|p{0.65\textwidth}|p{0.35\textwidth}|}
\hline
Moving {[}something{]} and {[}something{]} away from each other                                                                                     & \multirow{4}{*}{Moving two objects relative to each other}                   \\
Moving {[}something{]} and {[}something{]} closer to each other                                                                                     &                                                                              \\
Moving {[}something{]} closer to {[}something{]}                                                                                                    &                                                                              \\
Moving {[}something{]} away from {[}something{]}                                                                                                    &                                                                              \\ \hline
Moving {[}part{]} of {[}something{]}                                                                                                                & \multirow{2}{*}{Moving/Touching a part of something}                         \\
Touching (without moving) {[}part{]} of {[}something{]}                                                                                             &                                                                              \\ \hline
Opening {[}something{]}                                                                                                                             & \multirow{4}{*}{Opening or closing something}                                \\
Pretending to close {[}something{]} without actually closing it                                                                                     &                                                                              \\
Pretending to open {[}something{]} without actually opening it                                                                                      &                                                                              \\
Closing {[}something{]}                                                                                                                                                                                                                                    &                                                                              \\ \hline
Picking {[}something{]} up                                                                                                                          & \multirow{2}{*}{Picking something up}                                        \\
Pretending to pick {[}something{]} up                                                                                                                                                                                                                                                                                                                                                  &                                                                              \\ \hline
Piling {[}something{]} up                                                                                                                           & Piles of stuff                                                               \\ \hline
\begin{tabular}[c]{@{}l@{}}Plugging {[}something{]} into {[}something{]} but pulling it right out as you remove \\ your hand\end{tabular}           & \multirow{2}{*}{Plugging something into something}                           \\
Plugging {[}something{]} into {[}something{]}                                                                                                       &                                                                              \\ \hline
Poking {[}something{]} so it slightly moves                                                                                                                                                                                                                & \multirow{9}{*}{Poking something}                                            \\
Poking {[}something{]} so lightly that it doesn't or almost doesn't move                                                                            &                                                                              \\
Poking a stack of {[}something{]} without the stack collapsing                                                                                      &                                                                              \\
Poking a hole into {[}something soft{]}                                                                                                             &                                                                              \\
Pretending to poke {[}something{]}                                                                                                                  &                                                                              \\
Poking {[}something{]} so that it falls over                                                                                                        &                                                                              \\
Poking a stack of {[}something{]} so the stack collapses                                                                                            &                                                                              \\
Poking a hole into {[}some substance{]}                                                                                                             &                                                                              \\
Poking {[}something{]} so that it spins around                                                                                                      &                                                                              \\ \hline
Trying to pour {[}something{]} into {[}something{]}, but missing so it spills next to it                                                            & \multirow{6}{*}{Pouring something}                                           \\
Pretending to pour {[}something{]} out of {[}something{]}, but {[}something{]} is empty                                                             &                                                                              \\
Pouring {[}something{]} out of {[}something{]}                                                                                                      &                                                                              \\
Pouring {[}something{]} onto {[}something{]}                                                                                                        &                                                                              \\
Pouring {[}something{]} into {[}something{]}                                                                                                        &                                                                              \\
Pouring {[}something{]} into {[}something{]} until it overflows                                                                                     &                                                                              \\ \hline
Pulling {[}something{]} from behind of {[}something{]}                                                                                              & \multirow{5}{*}{Pulling something}                                           \\
Pulling {[}something{]} from right to left                                                                                                          &                                                                              \\
Pulling {[}something{]} out of {[}something{]}                                                                                                      &                                                                              \\
Pulling {[}something{]} onto {[}something{]}                                                                                                        &                                                                              \\
Pulling {[}something{]} from left to right                                                                                                          &                                                                              \\ \hline
Pulling two ends of {[}something{]} but nothing happens                                                                                             & \multirow{3}{*}{Pulling two ends of something}                               \\
Pulling two ends of {[}something{]} so that it separates into two pieces                                                                            &                                                                              \\
Pulling two ends of {[}something{]} so that it gets stretched                                                                                       &                                                                              \\ \hline
Pushing {[}something{]} onto {[}something{]}                                                                                                        & \multirow{9}{*}{Pushing something}                                           \\
Pushing {[}something{]} from right to left                                                                                                          &                                                                              \\
Pushing {[}something{]} with {[}something{]}                                                                                                        &                                                                              \\
Pushing {[}something{]} so that it falls off the table                                                                                              &                                                                              \\
Pushing {[}something{]} so that it almost falls off                                                                                                 &                                                                              \\
Pushing {[}something{]} off of {[}something{]}                                                                                                      &                                                                              \\
Pushing {[}something{]} so that it slightly moves                                                                                                   &                                                                              \\
Pushing {[}something{]} from left to right                                                                                                          &                                                                              \\ \hline
Pretending to put {[}something{]} on a surface                                                                                                                                                                                                                                                                                                                     & \multirow{2}{*}{Putting something somewhere}                                 \\
Putting {[}something{]} on a surface                                                                                                                                                                                                                                                                                                                                                                                                                                    &                                                                              \\ \hline
Laying {[}something{]} on the table on its side, not upright                                                                                        & \multirow{3}{*}{Putting something upright/on its side}                       \\
\begin{tabular}[c]{@{}l@{}}Putting {[}something that cannot actually stand upright{]} upright on the table, so it \\ falls on its side\end{tabular} &                                                                              \\
Putting {[}something{]} upright on the table                                                                                                        &                                                                              \\ \hline
\end{tabular}
\end{table}
\begin{table}[H]
\begin{tabular}{|p{0.65\textwidth}|p{0.35\textwidth}|}
\hline

Putting {[}something{]} underneath {[}something{]}                                                                                                  & \multirow{19}{*}{\parbox{6cm}{Putting/Taking objects into/out of/next to/… other objects}} \\
Putting {[}something{]} onto {[}something else that cannot support it{]} so it falls down                                                           &                                                                              \\
Failing to put {[}something{]} into {[}something{]} because {[}something{]} does not fit                                                            &                                                                              \\
Putting {[}something{]}, {[}something{]} and {[}something{]} on the table                                                                         &                                                                              \\
Pretending to put {[}something{]} behind {[}something{]}                                                                                            &                                                                              \\
Putting {[}something{]} in front of {[}something{]}                                                                                                 &                                                                              \\
Taking {[}something{]} out of {[}something{]}                                                                                                       &                                                                              \\
Pretending to put {[}something{]} onto {[}something{]}                                                                                              &                                                                              \\
Putting {[}something{]} and {[}something{]} on the table                                                                                            &                                                                              \\
Pretending to take {[}something{]} out of {[}something{]}                                                                                           &                                                                              \\
Putting {[}something{]} onto {[}something{]}                                                                                                        &                                                                              \\
Pretending to put {[}something{]} into {[}something{]}                                                                                              &                                                                              \\
Pretending to put {[}something{]} underneath {[}something{]}                                                                                        &                                                                              \\
Putting {[}something{]} next to {[}something{]}                                                                                                     &                                                                              \\
Putting {[}something{]} behind {[}something{]}                                                                                                      &                                                                              \\
\begin{tabular}[c]{@{}l@{}}Putting {[}something{]} on the edge of {[}something{]} so it is not supported and falls \\ down\end{tabular}             &                                                                              \\
Removing {[}something{]}, revealing {[}something{]} behind                                                                                          &                                                                              \\
Pretending to put {[}something{]} next to {[}something{]}                                                                                           &                                                                              \\
Putting {[}something{]} into {[}something{]}                                                                                                        &                                                                              \\ \hline
Letting {[}something{]} roll along a flat surface                                                                                                   & \multirow{10}{*}{Rolling and sliding something}                              \\
Rolling {[}something{]} on a flat surface                                                                                                           &                                                                              \\
Putting {[}something{]} that can't roll onto a slanted surface, so it slides down                                                                   &                                                                              \\
Lifting a surface with {[}something{]} on it until it starts sliding down                                                                           &                                                                              \\
Letting {[}something{]} roll down a slanted surface                                                                                                 &                                                                              \\
Lifting a surface with {[}something{]} on it but not enough for it to slide down                                                                    &                                                                              \\
Letting {[}something{]} roll up a slanted surface, so it rolls back down                                                                            &                                                                              \\
Putting {[}something{]} onto a slanted surface but it doesn't glide down                                                                            &                                                                              \\
Putting {[}something{]} that can't roll onto a slanted surface, so it stays where it is                                                             &                                                                              \\
Putting {[}something{]} on a flat surface without letting it roll                                                                                   &                                                                              \\ \hline
Pretending to scoop {[}something{]} up with {[}something{]}                                                                                         & \multirow{2}{*}{Scooping something up}                                       \\
Scooping {[}something{]} up with {[}something{]}                                                                                                    &                                                                              \\ \hline
Showing {[}something{]} to the camera                                                                                                                                                                                                                                                                                                                                                                                                                                   & \multirow{2}{*}{Showing objects and photos of objects}                       \\
Showing a photo of {[}something{]} to the camera                                                                                                                                                                                                                                                                                                                 &                                                                              \\ \hline
Showing that {[}something{]} is empty                                                                                                                                                                                                                                                                                                                                                                                                                                   & \multirow{2}{*}{Showing that something is full/empty}                        \\
Showing that {[}something{]} is inside {[}something{]}                                                                                              &                                                                              \\ \hline
Moving {[}something{]} across a surface without it falling down                                                                                     & \multirow{2}{*}{Something (not) falling over an edge}                        \\
Moving {[}something{]} across a surface until it falls down                                                                                         &                                                                              \\ \hline
{[}Something{]} falling like a feather or paper                                                                                                     & \multirow{2}{*}{Something falling}                                           \\
{[}Something{]} falling like a rock                                                                                                                 &                                                                              \\ \hline
Moving {[}something{]} and {[}something{]} so they collide with each other                                                                          & \multirow{2}{*}{Something passing/hitting another thing}                     \\
Moving {[}something{]} and {[}something{]} so they pass each other                                                                                  &                                                                              \\ \hline
Spilling {[}something{]} next to {[}something{]}                                                                                                    & \multirow{3}{*}{Spilling something}                                          \\
Spilling {[}something{]} onto {[}something{]}                                                                                                       &                                                                              \\
Spilling {[}something{]} behind {[}something{]}                                                                                                     &                                                                              \\ \hline
Spinning {[}something{]} so it continues spinning                                                                                                   & \multirow{3}{*}{Spinning something}                                          \\
Spinning {[}something{]} that quickly stops spinning                                                                                                &                                                                              \\
Pushing {[}something{]} so it spins                                                                                                                 &                                                                              \\ \hline
Spreading {[}something{]} onto {[}something{]}                                                                                                      & \multirow{2}{*}{Spreading something onto something}                          \\
Pretending to spread 'air' onto {[}something{]}                                                                                                     &                                                                              \\ \hline
Pretending to sprinkle 'air' onto {[}something{]}                                                                                                   & \multirow{2}{*}{Sprinkling something onto something}                         \\
Sprinkling {[}something{]} onto {[}something{]}                                                                                                     &                                                                              \\ \hline
Squeezing {[}something{]}                                                                                                                           & \multirow{2}{*}{Squeezing something}                                         \\
Pretending to squeeze {[}something{]}                                                                                                               &                                                                              \\ \hline
\end{tabular}
\end{table}
\begin{table}[H]
\begin{tabular}{|p{0.65\textwidth}|p{0.35\textwidth}|}
\hline

Stacking {[}number of{]} {[}something{]}                                                                                                            & \multirow{2}{*}{Stacking or placing N things}                                \\
Putting {[}number of{]} {[}something{]} onto {[}something{]}                                                                                        &                                                                              \\ \hline
Stuffing {[}something{]} into {[}something{]}                                                                                                       & Stuffing/Taking out                                                          \\ \hline
Taking {[}something{]} from {[}somewhere{]}                                                                                                                                                                                                                                                                                                                      & \multirow{2}{*}{Taking something}                                            \\
Pretending to take {[}something{]} from {[}somewhere{]}                                                                                             &                                                                              \\ \hline
Tearing {[}something{]} just a little bit                                                                                                           & \multirow{3}{*}{Tearing something}                                           \\
Pretending to be tearing {[}something that is not tearable{]}                                                                                       &                                                                              \\
Tearing {[}something{]} into two pieces                                                                                                             &                                                                              \\ \hline
Throwing {[}something{]} against {[}something{]}                                                                                                    & \multirow{6}{*}{Throwing something}                                          \\
Throwing {[}something{]}                                                                                                                            &                                                                              \\
Throwing {[}something{]} in the air and catching it                                                                                                 &                                                                              \\
Pretending to throw {[}something{]}                                                                                                                 &                                                                              \\
Throwing {[}something{]} in the air and letting it fall                                                                                             &                                                                              \\
Throwing {[}something{]} onto a surface                                                                                                             &                                                                              \\ \hline
\begin{tabular}[c]{@{}l@{}}Tipping {[}something{]} with {[}something in it{]} over, so {[}something in it{]} falls out\end{tabular}              & \multirow{2}{*}{Tipping something over}                                      \\
Tipping {[}something{]} over                                                                                                                        &                                                                              \\ \hline
Pretending to turn {[}something{]} upside down                                                                                                                                                                                                                      & \multirow{2}{*}{Turning something upside down}                               \\
Turning {[}something{]} upside down                                                                                                                                                                                                                                                                                                                              &                                                                              \\ \hline
Twisting (wringing) {[}something{]} wet until water comes out                                                                                       & \multirow{3}{*}{Twisting something}                                          \\
Pretending or trying and failing to twist {[}something{]}                                                                                           &                                                                              \\
Twisting {[}something{]}                                                                                                                            &                                                                              \\ \hline
Pretending or failing to wipe {[}something{]} off of {[}something{]}                                                                                & \multirow{2}{*}{Wiping something off of something}                           \\
Wiping {[}something{]} off of {[}something{]}                                                                                                       &                                                                              \\ \hline
\end{tabular}
\end{table}

\subsection{Data Collection Platform}

\begin{figure}[h]
\centering
\includegraphics[width=1.0\textwidth]{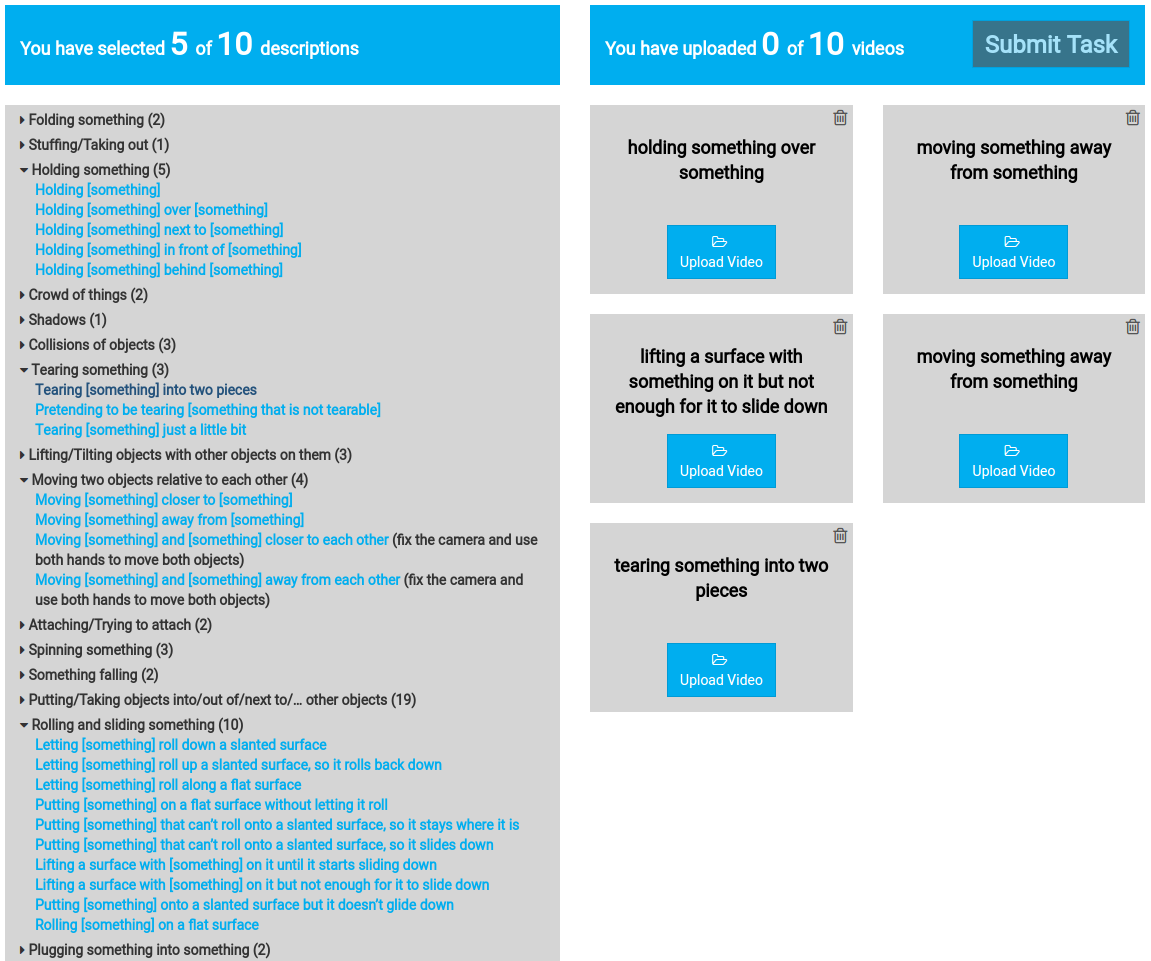}
\caption{Left Column: Crowd-workers can choose the classes they want to generate videos for (often the available classes are controlled so as to maintain class balance as much as possible). Right Column: An interface to upload videos and enter input-text.}
\label{figure:platformscreenshot1}
\end{figure}

\begin{figure}[h]
\centering
\includegraphics[width=1.0\textwidth]{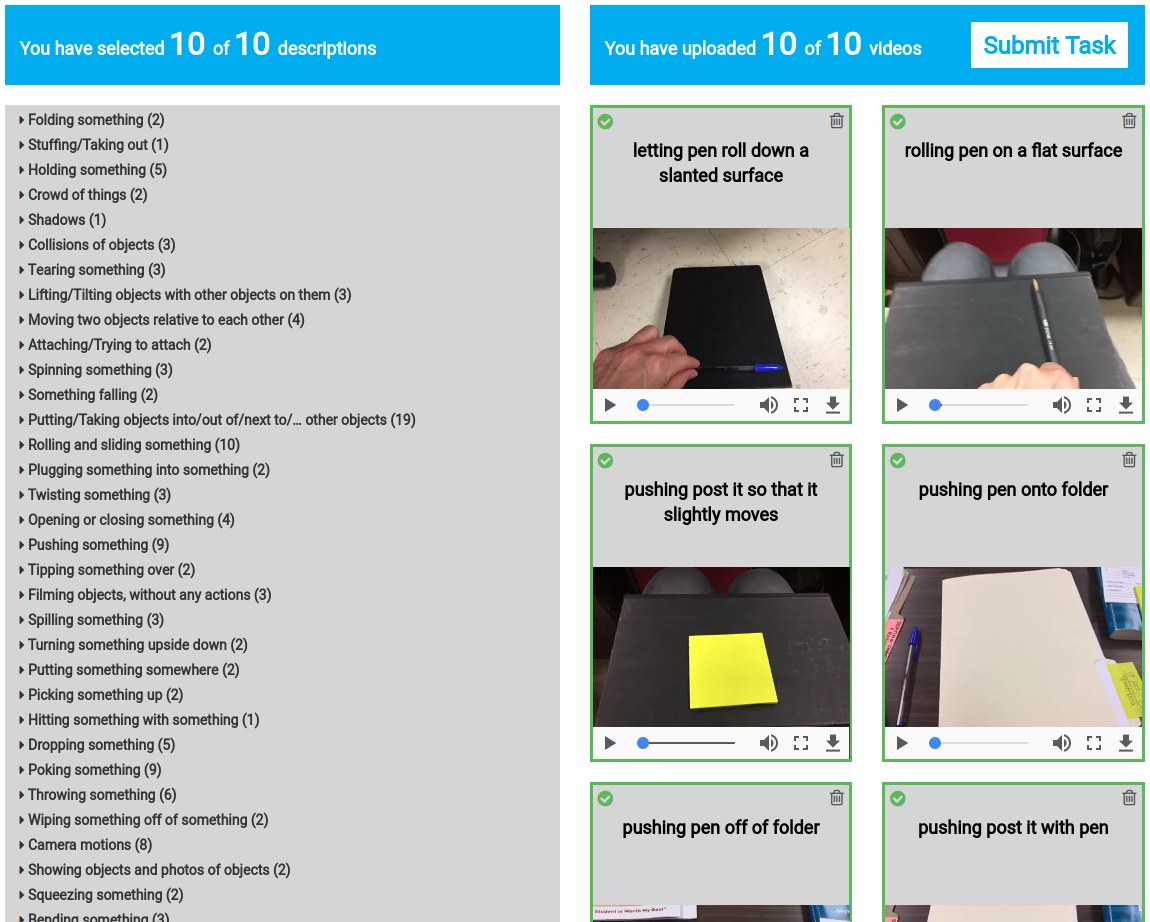}
\caption{An example of the upload interface after uploading videos and entering input-text.}
\label{figure:platformscreenshot2}
\end{figure}

We show some snapshots of our data collection platform in Figures~\ref{figure:platformscreenshot1} and \ref{figure:platformscreenshot2}. They demonstrate the platform used by the crowd-workers to select classes and to upload corresponding videos.

\end{document}